\documentclass[conference]{IEEEtran}

\newif\ifreview

\unless\ifreview
    \IEEEoverridecommandlockouts
\fi

\usepackage{cite}
\usepackage{amsmath,amssymb,amsfonts}
\usepackage{algorithmic}
\usepackage{graphicx}
\usepackage{textcomp}
\usepackage{xcolor}
\def\BibTeX{{\rm B\kern-.05em{\sc i\kern-.025em b}\kern-.08em
    T\kern-.1667em\lower.7ex\hbox{E}\kern-.125emX}}

\usepackage{paralist}
\usepackage{capt-of}
\usepackage{csquotes}
\usepackage{hyperref}
\usepackage{multirow}
\usepackage{enumitem}
\usepackage{tikz}

\usetikzlibrary{positioning,arrows.meta,fit}

\begin{document}

\title{Radial Basis Function Networks as Projection Heads in Self-Supervised Learning
\thanks{This work was funded by the Lower Saxony Ministry of Science and Culture as well as the Volkswagen Foundation as part of the research project \enquote{RLA - KI Reallabor Agrar} under grant number ZN4530.}
}

\ifreview
    \author{\IEEEauthorblockN{Anonymous}
    \IEEEauthorblockA{
    \textit{Anonymous Institute}\\
    \textit{ABC University}\\
    anonymous@abc.edu}}
\else
\author{
    \IEEEauthorblockN{
    Andreas Schliebitz and Heiko Tapken
    \IEEEauthorblockA{
    \textit{Faculty of Engineering and Computer Science}\\
    \textit{Osnabrück University of Applied Sciences}\\
    Osnabrück, Germany\\
    a.schliebitz@hs-osnabrueck.de\\
    h.tapken@hs-osnabrueck.de}}
    \and
    \IEEEauthorblockN{Martin Atzmueller}
    \IEEEauthorblockA{
    \textit{Semantic Information Systems Group}\\
    \textit{Osnabrück University}, and\\
    \textit{German Research Center for AI (DFKI)}\\
    Osnabrück, Germany\\
    martin.atzmueller@uos.de}}
\fi 
\maketitle
\begin{abstract}
Self-supervised learning (SSL) typically relies on a backbone encoder followed by a small multilayer perceptron (MLP) projection head, which is conventionally discarded after training, while backbone quality is assessed via costly linear probing on labeled data. We argue that this approach including discarding the projector is rather computationally wasteful. Instead, we propose replacing the MLP head with a radial basis function network (RBFN), whose interpretable center and shape parameters can be exploited to judge representation quality without labels or a separate classifier. To this end, we introduce Scale-Normalized Separation (SNS), a novel label-free quality metric derived solely from the kernel centers and shapes learned during training. Across five canonical SSL architectures (MoCo, SimCLR, BYOL, SwAV and SimSiam) and four image classification datasets, we show that RBFN projection heads are competitive drop-in replacements for standard MLP projectors. We recommend constructing them with three RBF layers activated by the Gaussian radial basis function. Moreover, SNS exhibits strong to very strong positive correlation with established logistic regression metrics, demonstrating that a trained RBFN projector can act as a reliable proxy for backbone representation quality. We additionally publish a novel PyTorch compatible image classification dataset based on Google's Open Images V7 to facilitate reproducible research into representation learning.
\end{abstract}

\begin{IEEEkeywords}
Nonlinear Projection Heads, Multilayer Perceptrons, Radial Basis Function Networks, Quality Metrics, Model Evaluation, Self-Supervised Learning, Representation Learning
\end{IEEEkeywords}

\section{Introduction\label{sec:introduction}}

Self-supervised learning (SSL) aims to learn meaningful representations from sparsely labeled data. Common SSL architectures~\cite{Chen2020a,Chen2020b,Grill2020,Caron2020,Chen2021} consist of two parts, the backbone $f(\cdot)$ and the projection head $g(\cdot)$. Backbones are usually deep convolutional neural networks (CNNs) or vision transformers (ViTs). Projection heads (also called projectors), implemented as shallow multilayer perceptrons (MLPs), have been shown in training to significantly improve the quality of learned representations~\cite{Chen2020b,Gupta2022,Xue2024}. Early experiments demonstrated that nonlinear projection heads, i.\,e. MLPs with a nonlinear activation function like ReLU, outperform linear projectors and can increase representation quality by up to 10\,\%~\cite{Chen2020b,Jing2022}, illustrating the importance of the projection head in SSL. Furthermore, there is not only a basic architectural consensus within the field of SSL but also on training, testing and inference protocols~\cite{Wittscher2026}. The entire network $g(f(\cdot))$ is trained end-to-end on the train split of a sufficiently large image dataset using gradient descent, adjusting its weights in accordance to some loss function applied to the outputs of the projector. After training however, the projector with its trained weights is typically discarded~\cite{Gupta2022,Chen2020b,Wittscher2026} and the backbone model is evaluated by training a linear classifier~\cite{Chen2020b,Grill2020} on the labeled test split. Once model quality is deemed sufficient in testing, a final transfer learning step is conducted to adapt the backbone's weights to a specific downstream task like image classification or segmentation~\cite{Wittscher2026}.

This paper is motivated by the observation that unconditionally discarding the projector after training appears wasteful not only from a computational but also information theoretical point of view. We demonstrate, how training a specific type of nonlinear projection head, namely a radial basis function network (RBFN)~\cite{Broomhead1988}, can be exploited to judge representation quality of the backbone without the need for a labeled test split and by extension a linear classifier.
The main contributions of this paper can be summarized as follows:
\begin{compactenum}
    \item We present an RBFN projection and prediction head as a drop-in replacement for most default MLP heads used in SSL, with full compatibility with the popular LightlySSL framework~\cite{Susmelj2020}.
    \item We propose a new quality metric called SNS (Scale-Normalized Separation) for scoring a backbone's representation quality based on center and shape parameters learned by the RBFN projector during training. We show through statistical correlation between our novel and current metrics that scoring representation quality is possible solely by interpreting those parameters.
    \item Furthermore, we present the efficacy of our RBFN projector and quality metric by conducting an empirical study on five different SSL architectures and four image classification datasets paired with a grid search over different radial basis functions, number of kernels, projector depths and RBFN normalizations. For that, we also publish a new PyTorch compatible \texttt{Dataset} implementation for representation learning based on subsets from Google's Open Images V7 dataset. 
\end{compactenum}

The results of this paper can be reproduced using the code available in the following GitHub repository: \ifreview\url{https://anonymous.4open.science/r/rbfn-ssl-projector}\else\url{https://github.com/andreas-schliebitz/rbfn-ssl-projector}\fi

\section{Related Work}\label{sec:related-work}

\subsection{Self-Supervised Visual Representation Learning}\label{sec:ssl-visual-representation-learning}

Contrastive and non-contrastive self-supervised learning (SSL) methods have rapidly closed the gap to supervised pretraining on standard image benchmarks. Contrastive approaches such as MoCo~\cite{Chen2020a} and SimCLR~\cite{Chen2020b} learn invariances by pulling together augmented views of the same image while pushing apart views of different images.
Non-contrastive methods avoid explicit negatives: BYOL~\cite{Grill2020} and SimSiam~\cite{Chen2021} rely on asymmetric predictor networks and stop-gradient operations, while SwAV~\cite{Caron2020} enforces consistency between cluster assignments of different views.

All of these frameworks share a common architectural pattern: a backbone encoder followed by a small multilayer perceptron projection head, which is discarded after pretraining. Our work targets this projection head directly, replacing it with a shallow radial basis function network across all five canonical frameworks.

\subsection{The Role of Projection and Prediction Heads}\label{sec:role-of-projection-and-prediction-heads}

Although projection heads were originally introduced as a minor implementation detail in SimCLR~\cite{Chen2020b}, subsequent analyses have established that they are crucial for downstream transfer. Chen~et~al.~\cite{Chen2020b} observed that linear probing on the projection-head input substantially outperforms probing on its output, motivating the practice of discarding the head after pretraining. Gupta~et~al.~\cite{Gupta2022} provide a theoretical account of this phenomenon, showing that the projection head absorbs invariances induced by the contrastive loss that would otherwise damage class-relevant features in the backbone. Bordes~et~al.~\cite{Bordes2023} systematically study this \enquote{guillotine regularization} effect, showing that the optimal cut-off layer depends on the alignment between pretext and downstream tasks.
Further work has examined the geometry of representations inside the head~\cite{Jing2022} and the impact of head depth and width~\cite{Xue2024}.

To the best of our knowledge, however, no prior work has investigated whether non-standard MLP heads with strong inductive biases, such as radial basis functions, can serve as drop-in replacements while preserving or even improving the quality of the underlying backbone.

\subsection{Radial Basis Function Networks in Deep Learning}

Radial basis function networks, introduced by Broomhead and Lowe~\cite{Broomhead1988} and analyzed by Park and Sandberg~\cite{Park1991} as universal approximators, were a mainstay of classical neural network research but have received comparatively little attention in the deep learning era. In the self-supervised setting, kernel-based perspectives on contrastive learning~\cite{Li2021} analyze established MLP-based losses through a kernel lens rather than introducing explicit RBF components into the architecture. In this paper we present, to the best of our knowledge, the first systematic integration of trainable RBFN heads into mainstream SSL pipelines.

\subsection{Label-Free Evaluation of Self-Supervised Representations}

The standard protocol for evaluating SSL backbones, i.\,e., linear probing with labeled data~\cite{Kornblith2019}, is computationally costly and fundamentally dependent on a downstream supervised dataset. This has motivated several label-free or proxy evaluation metrics, most prominently RankMe~\cite{Garrido2023}, a measure of the effective rank of learned features that correlates with linear-probe accuracy without requiring labels. Thilak~et~al.~\cite{Thilak2024} extend this idea with LiDAR, a discriminative variant that incorporates augmentation structure.

Our proposed metric differs from those mentioned above by deriving its signal from a trained head rather than from the backbone features directly: the learned centers and shapes of the Gaussian RBF encode information about cluster structure in the feature space that, as we show empirically, tracks downstream accuracy as measured by evaluating a logistic regression model on features generated by the trained backbone.

\section{Method}\label{sec:method}

\subsection{Preliminaries and Idea}

A radial basis function network is an artificial neural network that is activated using radial basis functions (RBFs), a family of mostly nonlinear real valued functions which map the distance $r=\parallel\mathbf{x}-\mathbf{c}\parallel$ of an input vector $\mathbf{x}\in\mathbb{R}^n$ to a fixed point $\mathbf{c}\in\mathbb{R}^n$ onto a scalar. If the fixed point differs from the origin, $\varphi_\mathbf{c}$ is called a radial kernel centered at $\mathbf{c}$. A well known and widely used RBF is the Gaussian, although others exist as discussed in Section \ref{sec:contributions}:
\begin{equation}
\varphi(r)=\exp(-(\varepsilon r)^2)
\label{eq:gauss-rbf}
\end{equation}
The output of an RBFN can be described as a linear combination of such radial basis functions applied to some input vector $\mathbf{x}\in\mathbb{R}^n$,
\begin{equation}
\phi\colon\mathbb{R}^n\to\mathbb{R},\,\mathbf{x}\mapsto\sum^N_{i=1}w_i\varphi_\mathbf{c}(\parallel\mathbf{x}-\mathbf{c}_i\parallel)
\label{eq:rbfn}
\end{equation}
where $N$ is the number of neurons in the hidden layer, $\mathbf{c}_i\in\mathbb{R}^n$ the center fixed point for neuron $i$ and $w_i\in\mathbb{R}$ the $i$-th output neuron's weight. If an RBF with a required shape parameter $\varepsilon$ is chosen, the $i$-th neuron will contain its own $\varepsilon_i$, completing the set $\{\mathbf{w},\varepsilon_i,\mathbf{c}_i\}$ of learnable parameters for that neuron.

Trained radial basis function networks are known to be highly explainable in comparison to regular neural networks through their use of radial basis functions as nonlinearities. Interpretability results from a special set of learnable parameters which mathematically describe the shape $\varepsilon$ and location $\mathbf{c}$ (centers) of the fitted RBFs within the network's projection space~\cite{Orr1996}. Since self-supervision is inherently based on feature similarity, methods like MoCo and SimCLR learn and retain clusters of visually similar embeddings in their backbone's latent space~\cite{Parulekar2023}. We hypothesize that the quality of such a clustering can be characterized using the aforementioned RBFN parameters since the backbone is optimized using the outputs of the projector. Assuming this hypothesis holds, representation quality of the backbone could be scored via some quality function calculated on the parameters learned by the RBFN projector during training. We therefore propose the following changes to established SSL architectures and evaluation protocols:

\begin{enumerate}
    \item Replacement of the generic MLP heads with our custom RBFN projection and prediction heads, which use highly interpretable radial basis functions, such as the Gaussian RBF, as their nonlinearities.
    \item Calculation of a quality metric which is solely based on the parameters learned by the RBFN projector during self-supervised training. During evaluation, this could replace both the training of a separate logistic regression (LogReg) model as well as the labeled test split.
\end{enumerate}

\subsection{Contributions}\label{sec:contributions}

\subsubsection{RBFN Projection Head}

We base our RBFN projection head (see Figure \ref{fig:rbfn-projection-head}) on a PyTorch compatible \texttt{nn.Module} implementation of a radial basis function layer developed by Russo~\cite{Russo2021}. This RBF layer introduces four new hyperparameters, namely the number of kernels ($K=128$)\footnote{Values given in parentheses are default values used in our implementation.} to use, whether the RBF outputs should be normalized or not (\texttt{true}) with $n\in\{\mathrm{true},\mathrm{false}\}$, the type of distance function $f_d$ (\texttt{euclidian}) being applied to $\mathbf{x}-\mathbf{c}_i$ and the radial basis function $f_b$ (\texttt{gaussian}) to process the resulting distance $r$. The number of effective parameters can immediately be reduced to three, since the distance function is almost always chosen to be Euclidean. Our experiments further indicate that normalization can also be disregarded in most cases, reducing the number of additional hyperparameters to only two when compared to generic MLP projectors. For activating the RBF layer within our projector, we provide ready-to-use implementations of the most common radial basis functions like the Gaussian RBF but also quadratic, multiquadric and spline variants. In our experiments, we choose two RBFs which are strictly positive definite (gaussian, inverse multiquadric) and two which are not (multiquadric, thin plate spine (TPS)). If the RBF is strictly positive definite, the resulting kernel matrix is guaranteed to be symmetric positive definite and therefore invertible~\cite{Micchelli1986}. This property ensures unique results for the weights of the RBFN while simultaneously stabilizing training by reducing sensitivity to noisy inputs. After training, the $K \times D$ matrix of $K$ kernel centers $c_1,\dots,c_K\in\mathbb{R}^D$ with $K$ associated shape parameters $\varepsilon_1,\dots,\varepsilon_K\in\mathbb{R}$ can be extracted from the RBF layer via predefined getter methods.

The RBFN projector does not use ReLU activation; instead, it relies on dedicated RBF layers to achieve nonlinearity while strictly adhering to the design principles of the LightlySSL framework. Our RBFN projector implements the framework's \texttt{ProjectionHead} interface allowing seamless interoperability with the selection of SSL architectures mentioned in Section \ref{sec:ssl-visual-representation-learning}. Note that, in the case of our RBFN projector, a \enquote{single layer} always refers to the combination of a linear layer, followed by optional batch normalization and an RBF layer acting as nonlinearity. The RBFN head itself is inspired by SimCLR's MLP projector~\cite{Chen2020b}, hence offering precise parametrization of input (2048), hidden (2048) and output dimensions (128), variable depth (3) as well as an optional batch normalization (\texttt{false}) step prior to the RBF layer. Since some SSL architectures like BYOL~\cite{Grill2020} and SimSiam~\cite{Chen2021} use an additional ReLU activated prediction head on top of the projector, we also provide our own implementation of a batch normalized RBFN prediction head (see Figure \ref{fig:rbfn-prediction-head}) consisting of a single RBF layer with the same default parameters as its projector counterpart. 

\begin{figure}[tbp]
    \centering
    \begin{minipage}[t]{0.49\columnwidth}
        \centering
        \begin{tikzpicture}[
            font=\footnotesize,
            >=latex,
            box/.style={
                draw,
                rounded corners,
                minimum width=3.8cm,
                minimum height=0.5cm,
                align=center
            },
            node distance=2mm
        ]
        
        \node[box, fill=gray!15] (in) {$\mathbf{h} \in \mathbb{R}^{2048}$};
        \node[box, fill=blue!10, below=of in] (l1) {\texttt{nn.Linear} $(2048\!\to\!2048)$};
        \node[box, fill=green!10, dashed, below=of l1] (bn0) {\texttt{nn.BatchNorm1d}};
        \node[box, fill=orange!20, below=of bn0] (r1) {\texttt{RBFLayer}($K,f_d,f_b,n$)};
        \node[below=2mm of r1] (d) {$\vdots$};
        \node[box, fill=blue!10, below=1mm of d] (lo) {\texttt{nn.Linear} $(2048\!\to\!128)$};
        \node[box, fill=green!10, dashed, below=of lo] (bn) {\texttt{nn.BatchNorm1d}};
        \node[box, fill=gray!15, below=of bn] (out) {$\mathbf{z} \in \mathbb{R}^{128}$};
        
        \foreach \a/\b in {in/l1,l1/bn0,bn0/r1,r1/d,d/lo,lo/bn,bn/out}
        \draw[->] (\a) -- (\b);
        \end{tikzpicture}
        
        \captionof{figure}{RBFN \emph{projection head} with stacked linear and RBF layers featuring RBF nonlinearities, preceded by an optional batch normalization step (disabled by default).}
        \label{fig:rbfn-projection-head}
    \end{minipage}
    \hfill
    \begin{minipage}[t]{0.49\columnwidth}
        \centering
        \begin{tikzpicture}[
            font=\footnotesize,
            >=latex,
            box/.style={
                draw,
                rounded corners,
                minimum width=3.8cm,
                minimum height=0.5cm,
                align=center
            },
            node distance=2mm
        ]
        
        \node[box, fill=gray!15] (in) {$\mathbf{z} \in \mathbb{R}^{128}$};
        \node[box, fill=blue!10, below=of in] (l1) {\texttt{nn.Linear} $(128\!\to\!2048)$};
        \node[box, fill=green!10, below=of l1] (bn0) {\texttt{nn.BatchNorm1d}};
        \node[box, fill=orange!20, below=of bn0] (r1) {\texttt{RBFLayer}($K,f_d,f_b,n$)};
        \node[box, fill=blue!10, below=of r1] (lo) {\texttt{nn.Linear} $(2048\!\to\!128)$};
        \node[box, fill=gray!15, below=of lo] (out) {$\mathbf{p} \in \mathbb{R}^{128}$};
        
        \foreach \a/\b in {in/l1,l1/bn0,bn0/r1,r1/lo,lo/out}
        \draw[->] (\a) -- (\b);
        \end{tikzpicture}
        \captionof{figure}{Batch-normalized RBFN \emph{prediction head} used by some architectures after the projector, consisting of a single RBF layer sandwiched between two linear layers.}
        \label{fig:rbfn-prediction-head}
    \end{minipage}
\end{figure}

\subsubsection{SNS Metric}

We propose a novel quality metric called Scale-Normalized Separation (SNS) which is calculated using the center and shape parameters learned by our RBFN projection head during self-supervised training. SNS is a simple and direct measure of kernel separation based on Gaussian interaction energy between kernel centers with scale-normalized distances. Let $\{c_k\}^K_{k=1}\subset\mathbb{R}^D$ denote the set of RBF kernel centers and $\{\varepsilon_k\}^K_{k=1}$ their associated kernel scales. For each pair of distinct kernels with $i \neq j$, we first calculate their dimension-corrected scale-normalized distance as

\begin{equation}
\delta_{i,j}=\frac{\parallel c_i-c_j\parallel_2}{\sqrt{D}\sqrt{\varepsilon_i^2+\varepsilon_j^2+\mu}}
\label{eq:scale-normalized-distance}
\end{equation}

where the factor $\sqrt{D}$ compensates for the growth of Euclidian distances in high-dimensional spaces and $\mu\in\mathbb{R}_{>0}$ is a small constant added for numerical stability. In order to quantify kernel separation based on these pairwise distances, we use the following Gaussian interaction function:

\begin{equation}
\psi:\mathbb{R}\to\mathbb{R},\,\delta_{i,j}\mapsto\delta^2_{i,j}\exp(-\delta^2_{i,j})
\label{eq:gaussian-interaction-function}
\end{equation}

Note that the energy of $\psi$ is maximized if $\delta_{i,j}=1$ but vanishes as $\delta_{i,j}$ approaches zero or infinity. Finally, the SNS score is defined as the average interaction energy between all unordered pairs of kernel centers represented by their scale-normalized distances:

\begin{equation}
\mathrm{SNS} = \frac{1}{K(K-1)}\sum_{\substack{i,j=1 \\ i \neq j}}^{K}\psi(\delta_{i,j})
\label{eq:sns}
\end{equation}

Our SNS metric is maximized when inter-center distances are well-matched to kernel scales, i.\,e. $\delta_{i,j}\approx1\;\forall i \neq j$. Further properties of SNS include dimension and scale invariance under joint rescaling of centers and shapes, symmetry due to a pairwise relation between kernel centers and full differentiability.

\subsubsection{Open Images V7 Subsets}

In order to simplify the exploration of model behavior in terms of class count, we publish~\ifreview\cite{Schliebitz2025_anon}\else\cite{Schliebitz2025}\fi~a new sampling mechanism for Google's Open Images V7 dataset (OI-V7)~\cite{Google2022} and provide a PyTorch compatible \texttt{Dataset} implementation with five pre-sampled subsets featuring 10, 30, 50, 70 and 100 classes. Our sampling strategy can be used to generate any balanced subset of the Open Images V7 dataset regardless of class count. However, sample counts per class can vary depending on the requested number of classes, since the smallest class within the original dataset automatically limits the size of all other classes in the subset. To reduce the practical downsides of this approach, we offer a built-in augmentation pipeline that can optionally augment each class to any size.

In addition to sampling flexibility, our Open Images V7 subsets also offer disjointness to the ImageNet-1K~\cite{Russakovsky2015} dataset, which is often used by popular machine learning libraries like Torchvision to pre-train architectures like ResNet~\cite{He2016}. Since weight initialization is widespread, practitioners have to be cautious when interpreting a model's fitting behavior on ImageNet-1K or subsets like ImageNet-100~\cite{Tian2020}; purely comparative settings such as ours are less affected. The third and last advantage of our Open Images V7 subsets over ImageNet-1K is mean sample resolution. On average, samples within our 100 class subset are approximately four times larger in pixel area ($964\times800$\,px)~\ifreview\cite{Schliebitz2025_anon}\else\cite{Schliebitz2025}\fi~compared to ImageNet-1K ($482\times415$\,px)~\cite{Russakovsky2013}. This property can be beneficial for augmentation heavy architectures like SimCLR, where Gaussian blur is often removed from the standard transformation pipeline when training with low resolution datasets like CIFAR100 ($32\times32$\,px) due to it visually distorting samples beyond recognition. Hence, with increasing image resolution, this and other kinds of heavy augmentation are less likely to compromise sample integrity.

\subsection{Experimental Setup}\label{sec:experimental-setup}

In this section, we conduct experiments in order to provide empirical evidence for the following assumptions:

\begin{enumerate}
    \item Replacing the default MLP projection head with a radial basis function network does not hurt model performance in any significant way.
    \item Representation quality can be judged using our SNS metric and the parameters learned by our RBFN projector during self-supervised training.
\end{enumerate}

We assume the first hypothesis to hold since suitably activated RBFN are proven~\cite{Park1991} to be universal approximators on a compact subset of $\mathbb{R}^n$. Hence, replacing a generic MLP projector with an equally capable RBFN should not impact performance in any meaningful way. However, it should be noted that RBFNs introduce a small set of additional hyperparameters which could either worsen or improve performance, depending on their specific values. The second hypothesis is assumed to be true since nonlinear optimization of the backbone is conducted using some loss function acting directly on the feature embeddings generated by the projector. Therefore, fitting explainable radial basis functions to those projections should provide qualitative insight into the much more complex embedding space learned by the backbone.  

\subsubsection{Establishing a Baseline}\label{sec:establishing-a-baseline}

In order to compare the efficacy of our RBFN projection heads with default MLP projectors, we first conduct baseline training and evaluation runs on the five well-known SSL architectures MoCo (v2)~\cite{Chen2020a}, SimCLR~\cite{Chen2020b}, BYOL~\cite{Grill2020}, SwAV~\cite{Caron2020} and SimSiam~\cite{Chen2021}. Baseline runs are carried out using the respective architecture's default projection and prediction head in combination with a ResNet-50~\cite{He2016} backbone pre-trained on ImageNet-1K~\cite{Russakovsky2015}. Each architecture is trained for 100 epochs on eight NVIDIA A100 80\,GB GPUs in distributed data parallel (DDP) mode using 
\ifreview\href{https://anonymous.4open.science/r/rbfn-ssl-projector/experiments/configs}{default hyperparameters}\else
\href{https://github.com/andreas-schliebitz/rbfn-ssl-projector/tree/master/experiments/configs}{default hyperparameters}\fi. For our baseline measurements, we vary only the depth of the default 2-layer MLP projector used by SimCLR and MoCo, as the implementations of BYOL, SwAV and SimSiam do not support this modification. The projector's depth is first increased to three and finally four layers where possible.

Since we conduct a purely comparative study, we use the well-established ImageNet-100 dataset~\cite{Tian2020} as our main benchmark. It consists of 100 randomly sampled classes from ImageNet-1K, with approximately 1300 samples per class. Model performance on fewer classes is explored via three image classification subsets generated from Google's Open Images V7 dataset with 10, 30 and 50 classes~\ifreview\cite{Schliebitz2025_anon}\else\cite{Schliebitz2025}\fi. Each dataset is subdivided into train, validation and test splits using a 70/10/20 split ratio. All samples are resized to $224\times224$\,px resolution before being normalized using the dataset's global per channel mean and standard deviation values.

After self-supervised training, we follow the standard protocol for model evaluation by fitting a logistic regression model to the feature vectors generated by the backbone on the labeled test split. The regression model is trained for 200 epochs with a batch size of 64 and a learning rate of $0.001$. We finally report accuracy ($A$), precision ($P$), recall ($R$) and $F_1$ score as metrics for the resulting linear classifier. In summary, we collect the results of nine training and evaluation runs for each of the four datasets, yielding a total of 36 baseline measurements.

\subsubsection{Training RBFN Projectors}\label{sec:training-rbfn-projectors}

For the training runs involving our RBFN projection heads, we use the exact same datasets and hyperparameters as with our baseline MLP projectors. The only major difference is the replacement of the default projection and prediction heads within each architecture with our custom RBFN-based implementations. Since our RBFN heads introduce a small set of new hyperparameters into the training process, we extend our baseline grid search to include not only projector depth but also the number of RBF kernels (128, 256, 512) and different radial basis functions (\texttt{gaussian}, \texttt{inverse\_multiquadric}, \texttt{multiquadric}, \texttt{tps}), used within either a standard or normalized RBFN architecture. After training, the backbone is evaluated using the same protocol and metrics as outlined in Section \ref{sec:establishing-a-baseline}. Due to an increased number of hyperparameters examined in our grid search, we conduct 216 different training runs on each dataset resulting in a total of 864 runs with training durations ranging from approximately 1.5 to 2.5 hours depending on the architecture. In Section \ref{sec:impact-of-rbfn-parameters}, we analyze the evaluation results using Shapley Additive Explanations (SHAP)~\cite{Lundberg2017} to derive evidence-based recommendations for suitable default values of the new hyperparameters.

\subsubsection{Model Evaluation using SNS\label{sec:model-evaluation-using-sns}}

Since SNS is computed from kernel centers and shapes, evaluation with this metric is only possible when an RBFN projection head is used during self-supervised training. Hence, the baseline runs described in Section \ref{sec:establishing-a-baseline} using a default MLP projector cannot be evaluated using this method. We instead show the efficacy of SNS by evaluating each RBFN run using both the well-known metrics calculated after logistic regression and SNS, followed by a correlation analysis quantifying the relationship between the two. We argue that high correlation is suitable to support our initial hypothesis that examining trained RBFN projectors via some quality function can provide valuable insights on the backbone's representation quality. After each of our RBFN training runs, we first evaluate the backbone using LogReg with established classification metrics. We additionally store the final model checkpoint, including the trained backbone and projection head, for further analysis. This procedure allows us to develop and test our SNS metric without having to retrain the entire projector from scratch every time we change our implementation.

In order to evaluate a run using SNS, we load the trained checkpoint's raw state dictionary and extract kernel centers and shapes from the deepest RBF layer within the RBFN projector:
\begin{itemize}
    \item If $s_{i,j}\in\mathbb{R}_{>0}$ denotes the SNS metric calculated for the $i$-th run on the $j$-th datasets with $1\leq i \leq 216$ and $1\leq j \leq 4$, then $\mathbf{S}\in\mathbb{R}_{>0}^{216\times4}$ is the resulting matrix of calculated SNS values for each run and dataset combination.
    \item Since we calculate four metrics (accuracy, precision, recall and $F_1$ score) on the LogReg model instead of only one, the correlation target is now a tensor $\mathbf{L}=(l_{i,k,j})\in\mathbb{R}^{216\times4\times4}$ with $l_{i,k,j}\in[0,1]\forall i,k,j$, where the second dimension indexes the aforementioned LogReg metrics and the last the four test datasets.
\end{itemize}
Hence, for each dataset $j\in\{1,\dots,4\}$ and each metric $k\in\{1,\dots,4\}$, we now consider the vectors
\begin{equation}
\begin{aligned}
\mathbf{s}_j &= (s_{i,j},\dots,s_{216,j})^\top \in \mathbb{R}_{>0}^{216},\\
\mathbf{l}_{k,j} &= (l_{1,k,j},\dots,l_{216,k,j})^\top \in \mathbb{R}_{[0,1]}^{216}
\end{aligned}
\end{equation}
where each $\mathbf{s}_j$ remains constant while being correlated two times with four different types of LogReg metrics per dataset. This is done by calculating Pearson's $r$~\cite{Pearson1895} and Spearman's $\rho$~\cite{Spearman1905} correlation coefficients between our SNS and the LogReg metrics as
\begin{equation}
r_{k,j}=\mathrm{corr}_r(\mathbf{s}_j,\mathbf{l}_{k,j}),\quad\rho_{k,j}=\mathrm{corr}_\rho(\mathbf{s}_j,\mathbf{l}_{k,j})
\end{equation}
yielding a pair of $4\times4$ correlation matrices, where the entry at position $(k,j)$ indicates the correlation between SNS and the $k$-th LogReg metric on the $j$-th test dataset. In contrast to $r\in[-1,1]$, the value of $\rho\in[-1,1]$ is a measure for any nonlinear monotonic relationship between two random variables or samples, whereas $r$ only measures the strength of their linear relationship. Additionally, Pearson's method also yields a $p$-value denoted $p_{k,j}$ testing the null hypothesis of no monotonic association between the inputs.

\section{Results\label{sec:results}}

\subsection{Projection Head Comparison\label{sec:projection-head-comparison}}

When comparing each architecture's default MLP projection head with our RBFN alternative, no significant performance difference can be observed. In Table \ref{tab:peak-backbone-performance}, we deliberately report the best evaluation runs based on accuracy, since all datasets exhibit nearly perfect inter-class balance. Although our RBFN head outperforms all MLP projectors on ImageNet-100 with up to +4\,\% accuracy on SimCLR and matches their performance on OpenImagesV7-10, it exhibits a modest performance deficit of 1--3\,\% on OpenImagesV7-50 and OpenImagesV7-30. However, our RBFN projector is particularly effective when used in conjunction with BYOL, where it manages to increase peak backbone performance by about 2-4\,\% across all datasets. On average, our RBFN projector is only outperformed by about half a percentage point in accuracy across different architectures and test datasets, which can be easily attributed to noise within our measurements.

\begin{table}[t]
\caption{Peak backbone performance by accuracy ($A$) using default MLP and our RBFN projection heads after LogReg evaluation.}
\label{tab:peak-backbone-performance}
\centering
\resizebox{\linewidth}{!}{
\setlength{\tabcolsep}{2pt}
\begin{tabular}{r|cccc|cccc|cccc|cccc}
\hline 
Dataset & \multicolumn{8}{c|}{\textbf{ImageNet-100}} & \multicolumn{8}{c}{\textbf{OpenImagesV7-50}}\tabularnewline
\hline 
Proj. head & \multicolumn{4}{c|}{\texttt{default}} & \multicolumn{4}{c|}{\texttt{rbfn}} & \multicolumn{4}{c|}{\texttt{default}} & \multicolumn{4}{c}{\texttt{rbfn}}\tabularnewline
\hline 
Metric & $F_{1}$ & $A$ & $P$ & $R$ & $F_{1}$ & $A$ & $P$ & $R$ & $F_{1}$ & $A$ & $P$ & $R$ & $F_{1}$ & $A$ & $P$ & $R$\tabularnewline
\hline 
MoCo & 0.92 & 0.92 & 0.92 & 0.92 & 0.92 & \underline{0.92} & \textbf{0.93} & 0.92 & \textbf{0.55} & \textbf{\underline{0.58}} & \textbf{0.56} & \textbf{0.58} & 0.53 & 0.56 & 0.55 & 0.55\tabularnewline
SimCLR & 0.88 & 0.88 & 0.89 & 0.88 & \textbf{0.92} & \textbf{\underline{0.92}} & \textbf{0.93} & \textbf{0.92} & \textbf{0.52} & 0.54 & 0.54 & 0.54 & 0.51 & \underline{0.54} & 0.54 & 0.54\tabularnewline
BYOL & 0.88 & 0.89 & 0.90 & 0.89 & \textbf{0.92} & \textbf{\underline{0.92}} & \textbf{0.93} & \textbf{0.92} & 0.49 & 0.52 & 0.55 & 0.52 & \textbf{0.53} & \textbf{\underline{0.55}} & \textbf{0.55} & \textbf{0.55}\tabularnewline
SwAV & 0.88 & 0.88 & 0.89 & 0.88 & \textbf{0.89} & \textbf{\underline{0.89}} & \textbf{0.90} & \textbf{0.89} & \textbf{0.47} & \textbf{\underline{0.51}} & \textbf{0.48} & \textbf{0.51} & 0.44 & 0.49 & 0.46 & 0.49\tabularnewline
SimSiam & 0.92 & 0.92 & 0.93 & 0.92 & 0.92 & \underline{0.92} & 0.93 & 0.92 & \textbf{0.55} & \textbf{\underline{0.57}} & \textbf{0.56} & \textbf{0.57} & 0.52 & 0.55 & 0.55 & 0.55\tabularnewline
\hline 
Average & 0.90 & 0.90 & 0.90 & 0.90 & \textbf{0.91} & \textbf{\underline{0.92}} & \textbf{0.92} & \textbf{0.92} & \textbf{0.52} & \textbf{\underline{0.55}} & \textbf{0.54} & \textbf{0.55} & 0.51 & 0.54 & 0.53 & 0.54\tabularnewline
\hline 
\hline 
Dataset & \multicolumn{8}{c|}{\textbf{OpenImagesV7-30}} & \multicolumn{8}{c}{\textbf{OpenImagesV7-10}}\tabularnewline
\hline 
Proj. Head & \multicolumn{4}{c|}{\texttt{default}} & \multicolumn{4}{c|}{\texttt{rbfn}} & \multicolumn{4}{c|}{\texttt{default}} & \multicolumn{4}{c}{\texttt{rbfn}}\tabularnewline
\hline 
Metric & $F_{1}$ & $A$ & $P$ & $R$ & $F_{1}$ & $A$ & $P$ & $R$ & $F_{1}$ & $A$ & $P$ & $R$ & $F_{1}$ & $A$ & $P$ & $R$\tabularnewline
\hline 
MoCo & \textbf{0.70} & \textbf{\underline{0.71}} & \textbf{0.71} & \textbf{0.71} & 0.65 & 0.67 & 0.68 & 0.67 & \textbf{0.76} & \textbf{\underline{0.76}} & \textbf{0.76} & \textbf{0.76} & 0.74 & 0.74 & 0.75 & 0.74\tabularnewline
SimCLR & \textbf{0.66} & \textbf{\underline{0.67}} & 0.67 & \textbf{0.67} & 0.64 & 0.66 & 0.67 & 0.66 & 0.73 & 0.73 & 0.73 & 0.73 & \textbf{0.74} & \textbf{\underline{0.74}} & \textbf{0.74} & \textbf{0.74}\tabularnewline
BYOL & 0.62 & 0.65 & \textbf{0.69} & 0.65 & \textbf{0.65} & \textbf{\underline{0.67}} & 0.68 & \textbf{0.67} & 0.72 & 0.72 & 0.72 & 0.72 & \textbf{0.74} & \textbf{\underline{0.74}} & \textbf{0.75} & \textbf{0.74}\tabularnewline
SwAV & \textbf{0.63} & \textbf{\underline{0.65}} & \textbf{0.66} & \textbf{0.65} & 0.61 & 0.63 & 0.64 & 0.63 & 0.70 & 0.70 & 0.70 & 0.70 & 0.70 & \underline{0.70} & 0.70 & 0.70\tabularnewline
SimSiam & \textbf{0.68} & \textbf{\underline{0.69}} & 0.68 & \textbf{0.69} & 0.65 & 0.67 & 0.68 & 0.67 & 0.73 & 0.73 & 0.73 & 0.73 & \textbf{0.74} & \textbf{\underline{0.74}} & \textbf{0.74} & \textbf{0.74}\tabularnewline
\hline 
Average & 0.66 & \textbf{\underline{0.67}} & \textbf{0.68} & \textbf{0.67} & 0.66 & 0.64 & 0.67 & 0.66 & 0.73 & 0.73 & 0.73 & 0.73 & 0.73 & \underline{0.73} & 0.73 & 0.73\tabularnewline
\hline 
\end{tabular}
}
\end{table}

\subsection{Impact of RBFN Parameters\label{sec:impact-of-rbfn-parameters}}

We conduct a SHAP analysis to systematically assess the effect of the RBFN parametrization described in Section \ref{sec:training-rbfn-projectors} on the backbone's prediction accuracy. Our results are summarized in Table \ref{tab:impact-of-rfbn-parameters} where $\overline{\phi^+_i}$ denotes the mean of the positive SHAP values and $\overline{\phi^-_i}$ the mean value of the negative SHAP values for hyperparameter $i$. If $\overline{\phi^+_i}>|\overline{\phi^-_i}|$, the hyperparameter exerts a net positive influence on prediction accuracy, while $\overline{|\phi_i|}$ quantifies the strength of its overall effect.

SHAP analysis indicates that RBFN projection heads should generally be activated using the Gaussian radial basis function. Our experiments also show that increasing the projector's depth to four layers without normalizing the RBFN's architecture allows the thin place spine to be used as an alternative activation function without compromising the backbone’s performance. Based on our experiments with SimCLR, BYOL and SwAV, we generally advise against using multiquadric and inverse multiquadric radial basis functions, as these can lead to significant performance losses and even trigger model collapse. In contrast to the radial basis function's profound impact on model performance, we do not record any significant fluctuations in backbone accuracy when varying the number of kernels fitted within each RBF layer. We therefore recommend adjusting the number of RBF kernels to match the number of compressed output features generated by the projector, which is typically 128.

With regards to the parameters that affect not only the RBF layers within the projector but the projector itself, our SHAP analysis confirms the generally observed benefit of increased projector depth~\cite{Chen2020c}. We find that slightly deeper projectors with three rather than two layers perform better on average, while even deeper RBFN projectors with four layers yield only marginal gains. Hence, we advise constructing RBFN projection heads with three instead of only two layers. Lastly, our SHAP results indicate, that RBFN normalization does not impact model performance in any significant way. However, as previously illustrated with TPS activation, some hyperparameter combinations might still require the RBF network to be normalized to achieve optimal performance. Since we do not find any clear advantages of this hyperparameter, we decide to eliminate it by advising against the use of normalization.

\begin{table}[t]
\caption{Impact of RBFN hyperparameters on backbone accuracy ($A$) after LogReg evaluation based on SHAP analysis.}
\label{tab:impact-of-rfbn-parameters}
\centering
\resizebox{\linewidth}{!}{
\setlength{\tabcolsep}{2pt}
\begin{tabular}{l|l|ccc|ccc|ccc|ccc}
\hline 
\multicolumn{2}{c|}{Dataset} & \multicolumn{3}{c|}{\textbf{ImageNet100}} & \multicolumn{3}{c|}{\textbf{OI-V7-50}} & \multicolumn{3}{c|}{\textbf{OI-V7-30}} & \multicolumn{3}{c}{\textbf{OI-V7-10}}\tabularnewline
\hline 
\multicolumn{2}{c|}{Mean SHAP} & $\overline{|\phi_{i}|}$ & $\overline{\phi_{i}^{+}}$ & $\overline{\phi_{i}^{-}}$ & $\overline{|\phi_{i}|}$ & $\overline{\phi_{i}^{+}}$ & $\overline{\phi_{i}^{-}}$ & $\overline{|\phi_{i}|}$ & $\overline{\phi_{i}^{+}}$ & $\overline{\phi_{i}^{-}}$ & $\overline{|\phi_{i}|}$ & $\overline{\phi_{i}^{+}}$ & $\overline{\phi_{i}^{-}}$\tabularnewline
\hline 
\multirow{5}{*}{RBF layer} & gaussian & 0.11 & 0.22 & -0.07 & 0.07 & 0.15 & -0.04 & 0.08 & 0.16 & -0.05 & 0.07 & 0.15 & -0.05\tabularnewline
 & TPS & 0.02 & 0.05 & -0.02 & 0.02 & 0.05 & -0.01 & 0.02 & 0.05 & -0.01 & 0.02 & 0.05 & -0.02\tabularnewline
 & mq. & 0.00 & 0.00 & -0.00 & 0.00 & 0.00 & -0.00 & 0.00 & 0.00 & -0.00 & 0.00 & 0.00 & -0.00\tabularnewline
 & mq.$^{-1}$ & 0.03 & 0.02 & -0.06 & 0.00 & 0.00 & -0.00 & 0.00 & 0.00 & -0.01 & 0.00 & 0.00 & -0.01\tabularnewline
\cline{2-14}
 & $\uparrow$ \#kernels & 0.02 & 0.02 & -0.02 & 0.01 & 0.01 & -0.01 & 0.02 & 0.01 & -0.03 & 0.01 & 0.01 & -0.01\tabularnewline
\hline
\multirow{2}{*}{RBFN head} & $\uparrow$ \#layers & 0.04 & 0.08 & -0.03 & 0.02 & 0.03 & -0.01 & 0.02 & 0.03 & -0.02 & 0.03 & 0.04 & -0.02\tabularnewline
 & normalize & 0.04 & 0.03 & -0.04 & 0.02 & 0.02 & -0.02 & 0.02 & 0.02 & -0.02 & 0.02 & 0.02 & -0.02\tabularnewline
\hline 
\end{tabular}
}
\end{table}

\subsection{Correlation of Classification Metrics with SNS}

After applying our SNS-based model evaluation protocol described in Section \ref{sec:model-evaluation-using-sns}, we compile our correlation results in Table \ref{tab:pearson-spearman-correlations}. The obtained correlation coefficients indicate a strong (0.60--0.79) to very strong (0.80--1.00) positive linear correlation between our SNS metric and all four established classification metrics computed after logistic regression. Additionally, all $p_{k,j}$-values obtained using Spearman’s rank correlation are close to zero, providing strong evidence against the null hypothesis and supporting a monotonic relationship between SNS and LogReg results.

\begin{table}[t]
\caption{Pearson ($r_{k,j}$) and Spearman ($\rho_{k,j}$) correlations of SNS with established performance indicators after logistic regression.}
\label{tab:pearson-spearman-correlations}
\centering
\resizebox{\linewidth}{!}{
\setlength{\tabcolsep}{2pt}
\begin{tabular}{r|cccc|cccc}
\hline 
Dataset ($j$) & \multicolumn{4}{c|}{\textbf{ImageNet-100}} & \multicolumn{4}{c}{\textbf{OpenImagesV7-50}}\tabularnewline
\hline 
Metric ($k$) & $F_{1}$ & $A$ & $P$ & $R$ & $F_{1}$ & $A$ & $P$ & $R$\tabularnewline
\hline 
$r_{k,j}$ & 0.87 & 0.87 & 0.85 & 0.87 & 0.93 & 0.92 & 0.92 & 0.92\tabularnewline
\hline 
$\rho_{k,j}$ & 0.85 & 0.85 & 0.85 & 0.85 & 0.83 & 0.83 & 0.69 & 0.83\tabularnewline
$p_{k,j}$ & $10^{-61}$ & $10^{-60}$ & $10^{-61}$ & $10^{-60}$ & $10^{-55}$ & $10^{-55}$ & $10^{-32}$ & $10^{-55}$\tabularnewline
\hline 
\hline 
Dataset ($j$) & \multicolumn{4}{c|}{\textbf{OpenImagesV7-30}} & \multicolumn{4}{c}{\textbf{OpenImagesV7-10}}\tabularnewline
\hline 
Metric ($k$) & $F_{1}$ & $A$ & $P$ & $R$ & $F_{1}$ & $A$ & $P$ & $R$\tabularnewline
\hline 
$r_{k,j}$ & 0.88 & 0.87 & 0.87 & 0.87 & 0.81 & 0.81 & 0.79 & 0.81\tabularnewline
\hline 
$\rho_{k,j}$ & 0.76 & 0.76 & 0.72 & 0.76 & 0.83 & 0.83 & 0.82 & 0.83\tabularnewline
$p_{k,j}$ & $10^{-41}$ & $10^{-41}$ & $10^{-34}$ & $10^{-41}$ & $10^{-56}$ & $10^{-57}$ & $10^{-52}$ & $10^{-57}$\tabularnewline
\hline 
\end{tabular}
}
\end{table}

\section{Discussion\label{sec:discussion}}

After analyzing the results presented in Table \ref{tab:peak-backbone-performance}, we come to the conclusion that shallow radial basis function networks with two to four RBF layers are a competitive replacement for most generic MLP projection heads used in self-supervised learning. This empirical observation supports our first hypothesis that replacing the default MLP with our RBFN projector does not significantly hurt performance. This outcome was expected, since RBFNs are proven universal approximators. Their hidden neurons correspond to localized kernel centers in the input space, making their decision process more transparent and interpretable than that of conventional MLPs. As a result, RBFNs can be regarded as a highly explainable type of multilayer perceptron that fits kernel centers to inputs using nonlinear RBF activation functions.

Once we have determined RBFNs to be a suitable alternative for MLP projectors, we continue to investigate their construction in terms of hyperparameter choice (see Table \ref{tab:impact-of-rfbn-parameters}). SHAP analysis suggests the Gaussian radial basis function to be the preferred method for activating our RBFN heads. Due to SSL being linked to clustering of feature vectors in latent space, our rationale for this finding is primarily based on the parameters learned by the RBFN during end-to-end training. We hypothesize that a multivariate normal distribution, being equivalent to a hyperdimensional Gaussian, provides the most flexible geometric shape for fitting clusters in feature space using shape ($\varepsilon_i$) and center ($\mathbf{c}_i$) parameters. In particular, the Gaussian RBF is highly localized due to its exponential decay as the distance $r$ between embeddings increases. The advantage of this property becomes obvious once applied to clustering, since points further away from center $\mathbf{c}_i$ should have reduced affinity and therefore a lower probability of belonging to the same cluster. Additionally, the learned shape parameters $\varepsilon_i$, which mathematically control the Gaussian's width, can be used to introduce the notion of neighborhood, essentially limiting the projections that should be included into the cluster centered at $\mathbf{c}_i$.

Further experiments on constructing our RBF layer with 128, 256 and 512 kernels show no significant impact on model performance. We do not find any evidence that the optimal number of RBF kernels is tied to the projector's output dimensions and instead suspect that even our lower bound of 128 kernels was already sufficient to fully saturate the projector. Verifying this hypothesis requires additional grid searches with a significantly reduced number of kernels.

Turning to the RBFN projection head itself, we observe a clear benefit from increasing the number of RBF layers from two to three. This phenomenon could be explained by the additional layer increasing the projector's buffering capacity through better isolation of the objective function from the outputs of the backbone, which are known to be less suitable for loss calculation. Instead, a third layer could allow the projections to distort more favorably in terms of the loss function, without hurting the valuable representations learned by the backbone. Additionally, deeper projectors can have a regularization-like effect on the backbone and could therefore prevent overfitting of the objective function. Lastly, we find normalization of our RBFN architecture to be ineffective, which could be explained by the fact that all samples were already normalized prior to training and testing using their associated dataset's mean and standard deviation values.   

Finally, we test our second hypothesis, which states that representation quality can be inferred from learned RBFN parameters. To this end, we evaluate each RBFN training run using our novel SNS metric and established linear classification metrics, followed by a correlation analysis of the resulting scores. Since both Pearson and Spearman correlation coefficients shown in Table \ref{tab:pearson-spearman-correlations} indicate a strong to very strong positive linear relationship between our SNS and LogReg metrics, we conclude that our second hypothesis also holds. Hence, a trained RBFN projector can be used as a proxy for judging backbone quality without the need for training a separate logistic regression model with labeled samples. We believe that correlation can be increased even further by either developing more sophisticated alternatives to our SNS metric or additional enhancements to our RBFN projector, improving the fit of kernel centers ($\mathbf{c}_i$) and shapes ($\varepsilon_i$) to the representations in projection space.

\bibliographystyle{IEEEtran}
\bibliography{bibliography}
\end{document}